\documentclass[sigconf,screen]{acmart}

\usepackage{microtype}
\usepackage{subfigure}
\usepackage{multirow} 
\usepackage{booktabs} 
\usepackage[utf8]{inputenc}
\usepackage[dvipsnames]{xcolor}
\usepackage[most]{tcolorbox}
\newcommand{\customfont}{}
\newcommand{\ptitle}[1]{\textit{#1}}
\usepackage[table]{xcolor}
\AtBeginDocument{%
  }

\setcopyright{acmlicensed}
\copyrightyear{2025}
\acmYear{2025}
\setcopyright{cc}
\setcctype{by}
\acmConference[MM '25]{Proceedings of the 33rd ACM International Conference on Multimedia}{October 27--31, 2025}{Dublin, Ireland}
\acmBooktitle{Proceedings of the 33rd ACM International Conference on Multimedia (MM '25), October 27--31, 2025, Dublin, Ireland}\acmDOI{10.1145/3746027.3755673}
\acmISBN{979-8-4007-2035-2/2025/10}

\acmSubmissionID{5179}

\begin{document}

\title{ArtRAG: Retrieval-Augmented Generation with Structured Context for Visual Art Understanding }

%
\author{Shuai Wang}
\email{s.wang3@uva.nl}
\affiliation{%
  \institution{University of Amsterdam}
  \city{Amsterdam}
  \country{NL}
}

\author{Ivona Najdenkoska}
\email{i.najdenkoska@uva.nl}
\affiliation{%
  \institution{University of Amsterdam}
  \city{Amsterdam}
  \country{NL}
}

\author{Hongyi Zhu}
\email{h.zhu@uva.nl}
\affiliation{%
  \institution{University of Amsterdam}
  \city{Amsterdam}
  \country{NL}
}

\author{Stevan Rudinac}
\email{s.rudinac@uva.nl}
\affiliation{%
  \institution{University of Amsterdam}
  \city{Amsterdam}
  \country{NL}
}

\author{Monika Kackovic}
\email{m.kackovic@uva.nl}
\affiliation{%
  \institution{University of Amsterdam}
  \city{Amsterdam}
  \country{NL}
}

\author{Nachoem Wijnberg}
\email{n.m.wijnberg@uva.nl}
\affiliation{%
  \institution{University of Amsterdam}
  \city{Amsterdam}
  \country{NL}
}
\additionalaffiliation{%
\institution{College of Business and Economics,
University of Johannesburg}
  \city{Johannesburg}
  \country{South Africa}
  }

\author{Marcel Worring}
\email{m.worring@uva.nl}
\affiliation{%
  \institution{University of Amsterdam}
  \city{Amsterdam}
  \country{NL}
}

\renewcommand{\shortauthors}{Shuai Wang et al.}


\begin{abstract}
Visual art understanding requires joint modeling of multiple perspectives and contextual inference rooted in cultural, historical, and stylistic knowledge. Recent multimodal large language models (MLLMs) demonstrate strong performance in generic captioning, primarily based on object recognition and training on large-scale generic data.  They struggle in providing captions incorporating the multiple perspectives that fine art demands. In this work, we introduce ArtRAG, a novel training-free framework that integrates structured knowledge into a retrieval-augmented generation (RAG) pipeline for multi-perspective artwork explanation.
ArtRAG automatically constructs an Art Context Knowledge Graph (ACKG) from domain-specific textual sources, organizing entities such as artists, themes, movements, and historical events into a rich, interpretable knowledge graph. At inference time, a multi-granular structured context retriever selects semantically and topologically relevant subgraphs to guide explanation generation. This approach enables MLLMs to produce contextually grounded, multi-perspective descriptions. Experiments on the SemArt and Artpedia datasets demonstrate that ArtRAG outperforms existing heavily trained baselines. Human evaluations further confirm ArtRAG’s ability to generate coherent, informative, and culturally enriched interpretations of artworks.  \noindent Our code base is available at: \url{https://github.com/ShuaiWang97/ArtRAG}.

\end{abstract}





\maketitle

%

\section{Introduction}


Understanding visual art goes beyond describing visible content—it requires interpreting the cultural, historical, and stylistic context embedded in the artwork. We define this broader interpretive task as visual art understanding, which we operationalize as context-aware painting explanation. This capability has important real-world applications, including assistive technology in museums and galleries, interactive educational platforms, and enriched online art exploration experiences.

Generating such meaningful explanations of visual artworks presents a unique challenge that extends far beyond traditional image captioning. The emergence of large foundation models, such as large language models (LLMs), has driven significant progress across a wide range of tasks in natural language processing and beyond~\cite{zhao2025surveylargelanguagemodels,brown2020, rag_review}. Recent advances in multimodal large language models (MLLMs), which extend LLMs with visual understanding capabilities~\cite{NEURIPS2023_llava, chen2023sharegpt4v, qwen2vl, zhang2025cchall, zhang2025vitcotvideotextinterleavedchainofthought, huang2025dual}, have demonstrated strong performance on general image description tasks.
However, they fall short in the domain of fine art due to their limited capability to incorporate historical, cultural, and stylistic context elements essential for producing human-like interpretations of artworks.

\begin{figure}[!t]
	\centering
	\includegraphics[width=\columnwidth,trim=34 10 20 0,  
        clip]{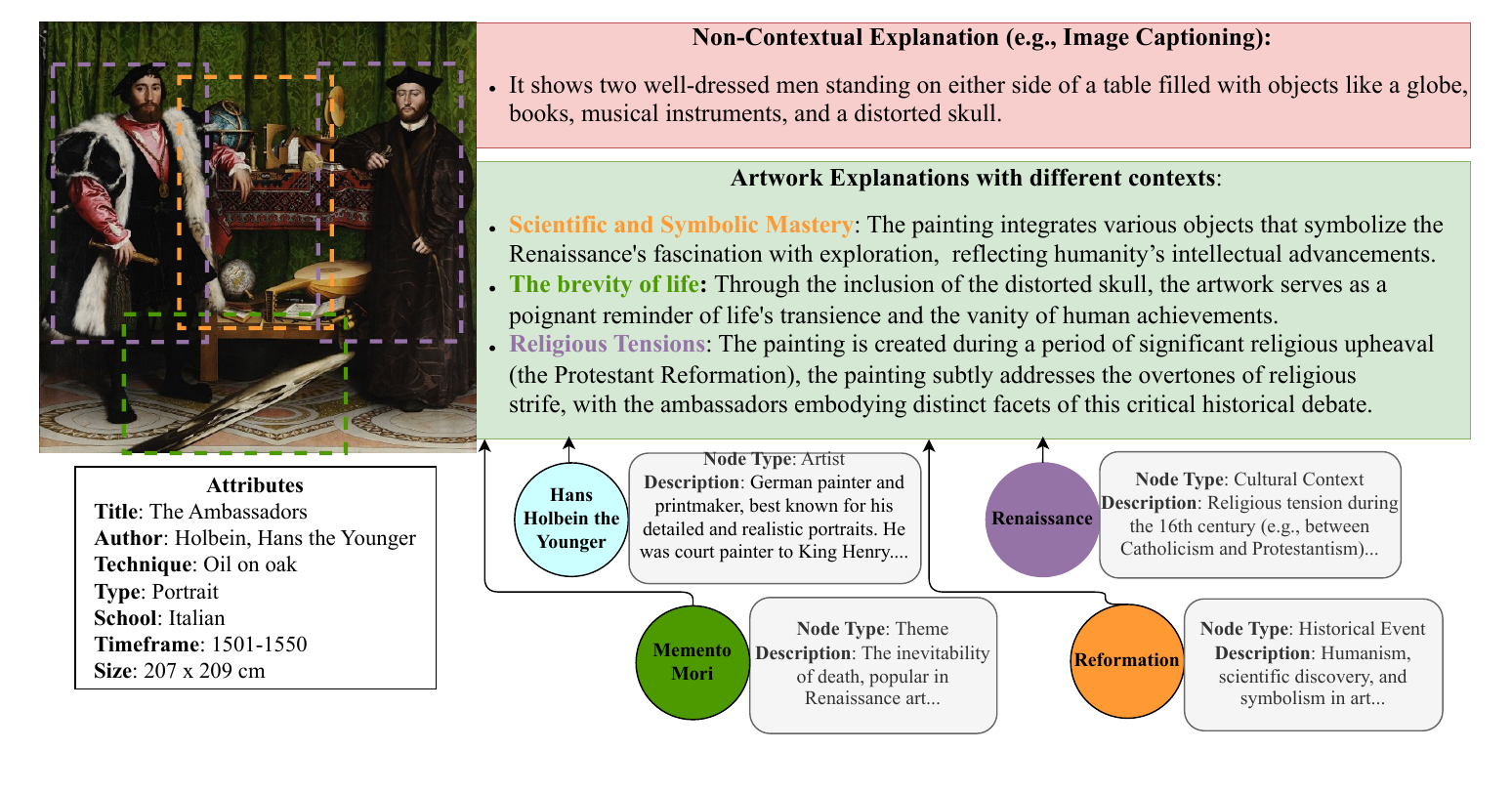}\vspace{-2mm}
	\caption{An example illustrating the interconnected contextual factors necessary for understanding \ptitle{The Ambassadors} by Hans Holbein the Younger. Explaining this painting requires not only information about the visual content (in red), but also a broader understanding across multiple contexts (in green): \textcolor{cyan}{artistic}, \textcolor{violet}{cultural}, \textcolor{ForestGreen}{thematic} and \textcolor{orange}{historical} context.}
    \vspace{-2mm}
    \label{fig:example}
\end{figure}


While existing methods attempt to generate artwork descriptions fine-tuning vision-language models on art datasets with structured metadata, they often rely on predefined schemas - such as artist, time period, or style - or limited ontologies that encode flat attributes like ``painted by'' or ``belonging to''~\cite{ijcai2024p848}. These representations fall short in enabling nuanced interpretive descriptions. In particular, they lack the ability to dynamically incorporate deeper contextual signals such as historical symbolism, cultural narratives, or artistic influence. Recent advances in  retrieval-augmented generation (RAG)~\cite{Bai2021ExplainMT} have shown promise in grounding open-ended generation tasks with external knowledge, but have yet to be effectively adapted to the art domain, where structured, multi-perspective context is essential for meaningful interpretation. As illustrated in Figure~\ref{fig:example}, explaining Hans Holbein the Younger’s \ptitle{The Ambassadors}\footnote{Titles of paintings are written in italics throughout this paper.} requires not only an understanding of the depicted scenes and objects, but also deeper reasoning across multiple dimensions with their explanations: artistic style (Northern Renaissance symbolism), historical context (Protestant Reformation), and philosophical themes (memento mori). These interconnected elements reflect the type of contextual reasoning that existing understanding systems and flat attribute-based knowledge graphs are unable to capture. 

To address these limitations, we introduce ArtRAG, a training-free retrieval-augmented generation framework that integrates structured art knowledge into the generative process. At the core of ArtRAG is the Art Context Knowledge Graph (ACKG), automatically constructed from art-related corpora to encode entities such as artists, movements, themes, techniques, and historical events, along with complex influence and context relations (e.g., “influenced by”, “depicts”, “emerged during”). These relations allow for tracing interpretive connections — e.g., linking an artist to a historical ideology or art school — thus enabling context-aware generation that reflects the underlying meaning embedded in visual artworks.
During inference time, given a painting and question, a multi-stage structured retriever selects a relevant subgraph from ACKG based on both textual and multimodal cues, enabling ArtRAG to generate visually-grounded and contextually-enriched descriptions that reflect multiple interpretive perspectives. 
 
To evaluate our method, we utilize the SemArt~\cite{semart_2018} and Artpedia~\cite{2019_Artpedia} datasets, which include manually created textual annotations for paintings. These annotations cover both visual content, and context dimensions. We demonstrate that our approach of incorporating structured contextual knowledge into RAG, significantly enhances the quality of generated descriptions compared to traditional RAG-based approaches and heavily trained baselines on these benchmarks. We also perform human evaluations to further confirm its ability to generate coherent, informative, and culturally-aware explanations.  Overall, our work bridges the gap between visual recognition and contextual understanding in artwork description, offering a robust framework for integrating structured information into generative models.  Our contributions are as follows:

\begin{itemize}

 \item We introduce \textbf{ArtRAG}, a novel training-free framework that integrates structured contextual retrieval with multimodal input for visual art understanding.

  \item We construct the \textbf{Art Context Knowledge Graph (ACKG)} from diverse art-related sources. ACKG captures rich, interconnected information beyond metadata, and will be released to support future research.

  \item We design a \textbf{multi-granularity structured retriever} that operates from coarse semantic similarity to fine-grained multimodal alignment, enabling highly relevant subgraph selection from ACKG for each artwork.


  \item ArtRAG achieves \textbf{superior results} compared to  heavily trained baseline methods across multiple evaluation benchmarks for art explanation without the need for any training, demonstrating its ability to produce interpretable, context-aware, and multi-perspective art descriptions.

\end{itemize}

\begin{figure*}[ht]
    \centering\includegraphics[width=\linewidth]{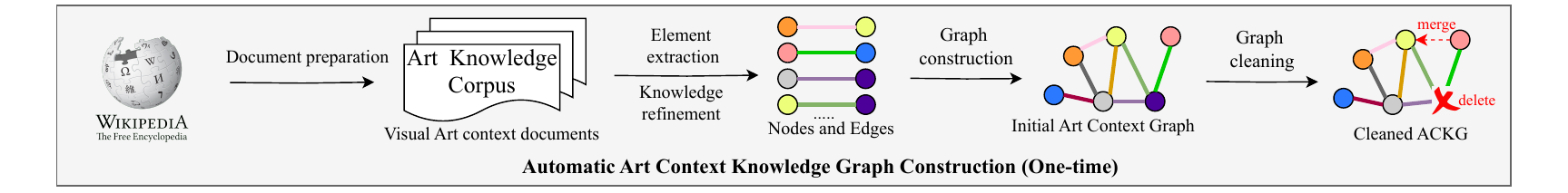}
    \caption{ Overview of the ArtRAG framework's automatic ACKG construction process. 
     Structured knowledge is extracted from visual art corpus through multi-perspective data collection and refinement. Artistic, historical, and thematic elements are identified and organized into an interconnected Art Context Knowledge Graph (ACKG). This structured representation enables efficient retrieval of contextual knowledge for paintings.}
    \label{fig:framework_a}

\end{figure*}

\section{Related work}
We discuss related work on two key subjects relevant to our approach: MLLMs for multimodal understanding and automatic artwork analysis.

\subsection{MLLMs for Multimodal Understanding}
In recent years, language modeling has evolved rapidly and achieved impressive progress~\cite{devlin_bert_2019, brown2020, chowdhery_palm_2022}. In the multimodal domain, vision language pre-training~\cite{10.1007/978-3-030-58577-8_7, clip_2021, yu2025crispsam2sam2crossmodalinteraction, huang2025ccsumsp} and MLLMs ~\cite{2023_BLIP, chen2023sharegpt4v, NEURIPS2023_llava, MiniGPT, sun_generative_2023} attempt to bridge visual perception and textual understanding, for enabling image-based description generation. However, these models primarily focus on object recognition and scene description, struggling with multiple perspectives of art context essential for fine art interpretation. Recent trends in visual art domain include, for example, finetuning of MLLMs using the visual art datasets, to enhance their art understanding ability on different perspectives like visual formal analysis~\cite{MM24GalleryGPT}, emotional analysis~\cite{Yuan2023ArtGPT4TA}, and 
general analysis~\cite{ExpArt_acl2024}. However, they rely on the internal knowledge of MLLMs and the finetuning dataset, which lacks factual knowledge like historical background and artist biography that are needed for comprehensive visual art understanding. Despite these successes, they inherit the intrinsic limitations of LLMs~\cite{Zhao_2024, zhao2025surveylargelanguagemodels}, such as the lack of domain-specific knowledge, the issue of “hallucination”, and the substantial computational
resources required for fine-tuning the model. RAG models, which integrate an external retrieval corpus to enhance factuality, have improved content generation~\cite{Zhao2023RetrievingMI, rag_survey,edge2025localglobalgraphrag,guo2024lightragsimplefastretrievalaugmented}. However, it is still hard to structure retrieved knowledge into meaningful, multi-perspective narratives.

\subsection{Automatic Artwork Analysis}
Recent advances in computer vision have addressed various art-related tasks such as painting classification~\cite{strezoski2020tindart, Art_ICIP_16, eccv12_art, 2017omniart, proto_HGNN}, retrieval~\cite{conde_clip-art_2021, BMVC_art_retrieval, Thanos_MM_21, semart_2018}, captioning~\cite{MM_19_art,Iconographic_IC, achlioptas2021artemis, mohamed_it_2022}, visual question answering~\cite{garcia_dataset_2020, bleidt_artquest_2024, ExpArt_acl2024} and painting generation~\cite{hu2024towards, Im2Oil_MM22}. However, these approaches typically focus on isolated aspects—such as visual content or stylistic features—without capturing multi-perspective art context critical for full interpretation. This limitation is largely due to the absence of structured, domain-specific knowledge representations.
Multi-perspective art description, which requires integrating visual analysis with contextual reasoning, remains underexplored. Prior works like Bai et al.~\cite{Bai2021ExplainMT} organize explanations into content, form, and context, and retrieve unstructured Wikipedia text. This lacks semantic structure and can lead to noisy or tangential descriptions KALE~\cite{ijcai2024p848} further leverages heterogeneous graphs to inject structured metadata, such as artist biographies and artistic movements, into captioning models. However, the above methods often rely on predefined taxonomies or static ontologies, limiting their ability to understand the broader context in which the artworks are created.
In contrast, ArtRAG constructs a domain-specific knowledge graph from raw art texts and employs a multi-stage structured retriever to extract subgraphs tailored to each painting. This enables generation of contextually grounded and comprehensive descriptions, bridging the gap between surface-level recognition and deeper contextual understanding.
\begin{figure*}[ht]
    \centering\includegraphics[width=\linewidth]{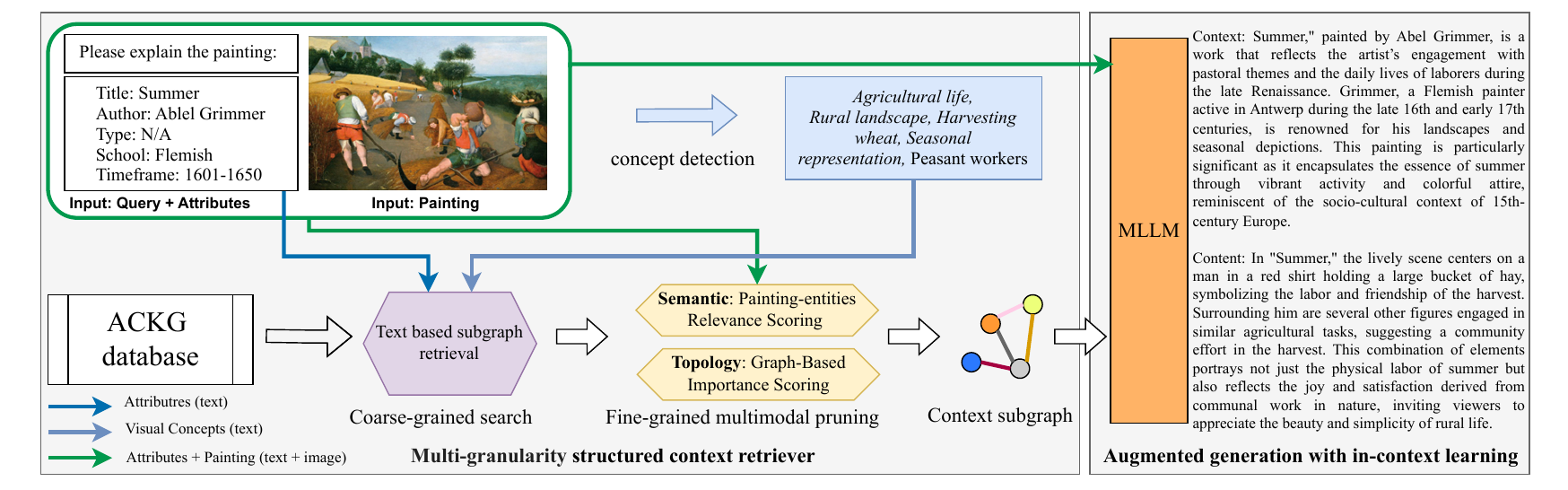}
    \vspace{-3mm}
    \caption{ Multi-stage retrieval and description generation in the ArtRAG framework. Given a painting (e.g., \ptitle{Summer by Abel Grimmer}), with its available attributes (title, artist, technique, etc.), the system first generates concepts. Then, a multi-granularity retrieval system retrieves the relevant sub-graph from the ACKG, leveraging both semantic and graph-based importance scoring. The retrieved structured knowledge is then integrated into an enhanced context subgraph, which guides the MLLM-based generation to produce detailed, contextualized painting descriptions.}
    \label{fig:framework_b}

\end{figure*}

\section{Preliminary}
\label{sec:Preliminary}
Before introducing our proposed approach, it is useful to first provide background knowledge on the conventional RAG framework~\cite{rag_review, zhao2024rag} to facilitate comparison with our method. The RAG framework consists of three main stages:

 ($1$) \textbf{Constructing of a knowledge database:} In the standard knowledge database-construction method, textual documents $x$ are chunked into small chunks ${x_n}$. These text chunks are then embedded into dense vector representations ${v_n}$ by a language model $L$. In addition, each embedding ${v_n}$ maintains an index that links it to its original text chunk ${x_n}$, allowing the retrieval of the original text content. These embeddings ${v_n}$ can be stored in a vector database $\mathcal{V}$ that facilitates efficient similarity search and retrieval.

 ($2$) \textbf{Knowledge retrieval:}  When the user inputs a query text $ \hat{x}_q $, it is embedded into a query representation $ \hat{v}_q $ using $L$. This query embedding is utilized to compute similarity scores across a vector database $ \mathcal{V} $, which contains all document chunk embeddings. The top-$k$  most similar embeddings $ \{\hat{v}_k\} = \{v_1, \ldots, v_k\} $, corresponding to document chunks $ \{\hat{x}_k\} = \{x_1, \ldots, , x_k\}$, are retrieved based on semantic proximity to $ \hat{v}_q $ in the vector space. These retrieved chunks $ \{\hat{x}_k\} $ serve as additional contextual information to augment the original user query.

 ($3$) \textbf{Answer generation:} The initial user query $ \hat{x}_q $ is concatenated with the retrieved document chunks $ \{\hat{x}_k\} $ to construct an expanded prompt $ p $ using a function $ \mathcal{P} $. Formally, $ p = \mathcal{P}(\hat{x}_q, \{\hat{x}_k\}) $. This enriched prompt $ p $, containing both the original query and the retrieved contextual knowledge, is then passed to an LLM. The LLM processes this input holistically, generating the final output $ z $ by analyzing the semantic relationships within $ p $.

  Our method enhances all three stages of the RAG framework for art analysis, and the details are given in the following section.

\section{Methodology}
ArtRAG consists of three stages linking to the three stages in Section~\ref{sec:Preliminary}: automatic art knowledge graph construction (one-time 
process) as illustrated in Figure~\ref{fig:framework_a}, multi-granularity structured context retrieval, and 
augmented generation with in-context learning shown in Figure~\ref{fig:framework_b}. In this section, we explain each stage separately.

\subsection{Art Context Knowledge Graph (ACKG) Construction}

As discussed in the introduction, effective interpretation of artworks requires knowledge that extends beyond visual attributes~\cite{belton1996art}, incorporating multi-perspective art context. However, existing textual information sources like books in the art domain are often noisy and unstructured, making it challenging for models to retrieve and synthesize relevant contextual knowledge. To address this, we construct an Art Context Knowledge Graph (ACKG) that organizes structured, interconnected knowledge for RAG. The construction process involves three key stages: document preparation, entity and relationship extraction, and art context graph cleaning.

\begin{table}[ht]
    \centering
    \caption{Summary statistics of the constructed ACKG. The table reports the number of nodes per type, the number of edges originating from those nodes, and the average length of their associated textual descriptions.
    }
    \label{tab:art_kg_stats}
    \begin{tabular}{lrrr}
        \toprule
        \textbf{Node Type} & \textbf{\# Nodes} & \textbf{\# Edges} & \textbf{Avg. Len} \\
        \midrule
        \textbf{Artists}       & 12,450  & 30,923 &  37.2 \\
        \textbf{Theme}     & 10,219 & 5,690  & 29.4 \\
        \textbf{Culture \& History} & 18,079    & 13,246   & 31.8\\
        \textbf{Art style  \& technique}    & 12,385  & 5,377  &  31.8 \\
        \textbf{Art Movement \& school}       & 2,502    & 6,603   & 41.3 \\
        \textbf{Others} & 767    & 118   & 18.4\\
        \midrule
        \textbf{Total (Overall Graph)} & 56,403 & 61,957  & 32.5\\
        \bottomrule
    \end{tabular}
\end{table}

\textbf{Document preparation:} To construct the ACKG, we automatically extract and structure knowledge from various textual sources in Wikipedia\footnote{https:/wikipedia-api.readthedocs.io} related to visual arts. We curate our knowledge based on Wikipedia by scraping and processing pages that provide relevant background information. However, Wikipedia is very large and often contains noisy and inaccurate or irrelevant content to the information need, necessitating careful selection and filtering. We extract entity pages associated with five key categories: Artists, Art Schools, Art Movements, Art Techniques, and Cultural Events.  The document preparation process is based on paining attributes used in  two art explanation generation datasets: SemArt~\cite{semart_2018}, Artpedia~\cite{2019_Artpedia} and a VQA dataset: ExpArt~\cite{ExpArt_acl2024}. Specifically:
SemArt provides predefined attribute tags, resulting in a collection of Wikipedia pages covering 410 artists, 20 art schools, and 9 art techniques, which we use as a starting point of our knowledge extraction process. Artpedia~\cite{2019_Artpedia} further contributes 180 additional artist pages, enriching the document source.  We also check ExpArt~\cite{ExpArt_acl2024}, which supplies relevant entities related to cultural events and art movements, enhancing the contextual understanding of artworks, using GPT4o~\cite{GPT-4o} to filter out important cultural events and art movements that are essential for artwork explanation.  In total, our ACKG construction results in 1,091 Wikipedia pages and a textual corpus comprising 2.76 million words, with more details in in Supplementary Materials. Notably, we do not include any wikipedia pages for paintings, which ensures our ACKG contains only contextual information. Our approach supports continuous expansion by incorporating new documents beyond Wikipedia, ensuring the knowledge graph remains scalable, up-to-date, and adaptable to emerging art context like trend.

Then we break down large, unstructured textual documents into smaller, more manageable passages, like a normal RAG system. This step is crucial for efficient processing and retrieval. By isolating smaller segments, our framework can quickly identify and extract relevant information without requiring exhaustive analysis of entire documents. Each segment is treated as a potential source of knowledge and is designed to capture both coarse-grained segments and fine-grained details such as high-level historical overviews, or specific biographical data about artists.

\textbf{Nodes and Relationships Extraction:}
Given the input text we want to extract factual and context dimensions of the input text to ensure the quality of the resulting knowledge graph. After the document preparation step is completed, we leverage LLMs to identify and extract entities and relationships from the text. These entities are categorized into predefined types relevant to the art domain. We utilize a series of prompts designed to guide the LLM in recognizing relevant information. For example, given a text that describes the influence of an artist, the model identifies the artist as an entity and links them to their movement, both with descriptions on entity and links. The relationships extracted during this stage form the edges of the graph, connecting nodes (entities) to reveal the interconnected nature of art history. To enhance the accuracy and domain specificity of node and relationship extraction, we design prompts for the art domain and use few-shot examples to guide the extraction, bypassing the need for additional training. More details in the supplementary materials.

\textbf{Art Context Graph cleaning:}
Due to redundancy among different sources and different expressions of the same entities, the initial set of nodes and edges extracted during the construction of the knowledge graph could have noise and inconsistency. To address this, we apply a refinement and synthesis process that merges semantically equivalent nodes and standardizes their associated descriptions.  Specifically, we perform fuzzy matching on node names to consolidate entities referring to the same information. 
We also ensure that the names with special characters, like Roman numerals in case of \ptitle{Elizabeth I} and \ptitle{Elizabeth II} are not merged. Additionally, we filter out nodes that fall outside the defined set of relevant entity types, retaining only those pertinent to our target schema (e.g., artists, themes, movements, techniques).

Before cleaning, the graph contains 58,208 nodes and 62,562 edges. After applying cleaning, the refined graph comprises 56,403 nodes and 61,957 edges with details in Table~\ref{tab:art_kg_stats}. This cleaned version forms a denser, more semantically consistent foundation for contextual retrieval and generation.

\subsection{Multi-granularity structured retriever}
 To generate contextually rich and informative descriptions for a given painting, we design a retrieval system that extracts relevant nodes and edges from an Art Knowledge Graph, forming a focused subgraph that provides essential contextual details. The retrieval process consists of two main stages: \textit{Coarse-Grained Subgraph Retrieval}, and \textit{Graph Pruning and Refinement}.

\textbf{Coarse-Grained Subgraph Retrieval:} The initial retrieval stage focuses on efficiently narrowing down the search space by retrieving a coarse subgraph based on textual embedding similarity. Given that paintings are inherently multimodal, directly comparing an image with graph-based knowledge is a challenge. To address this, we adopt a text-based approach at this stage as follows.

We first transform the painting image into a set of concise, Art domain concepts that serve as semantic cues for querying the knowledge graph. This transformation is performed using a small vision-language model (i.e., LLaVA-1.5), which analyzes the visual content and generates a list containing a fixed number of descriptors summarizing key aspects of the painting. For example, for the painting \ptitle{Summer} by Abel Grimmer in Figure~\ref{fig:framework_b}, the generated concepts, among others, may include ``Agricultural life''  and ``Rural landscape''. They act as the bridge between the visual modality and the textual knowledge graph, allowing the retrieval process to identify relevant entities aligned with the content and context of the painting. Together with the painting’s textual attributes (e.g., title, artist, time frame), these cues are encoded and used to retrieve the top-$k$ most relevant nodes from the ACKG based on text similarity in the shared embedding space.

\begin{equation}
\mathcal{V}_{\text{sub}} = \operatorname*{arg\ top\text{-} k'}_{v \in  V} \cos \left(z_p, z_v \right)
\end{equation}
where $z_p$ is the joint embedding of the painting attributes and extracted concepts, and $z_n$ is the embedding of node $v_n$ in the graph.
We then expand the graph by exploring important edges linked to the top-$k'$ retrieved nodes. To include contextually rich entities, we prioritize edges connecting to highly relevant or central nodes, using edge degree $|N(u) \cup N(v)| - 2$ as a measure. This expansion goes to top-$k$ nodes captures thematic links, such as shared artistic styles, movements, or influential figures.

\begin{table*}[ht]
\centering
\caption{Evaluation results on datasets Artpedia and SemArt over baselines. Bold represents the highest score. As we directly use the reported results for the competitors, some of the metrics are missing, we use ``--'' to indicate missing values.}
\label{tab:evaluation_metrics}

\begin{tabular}{llcccccccc}
\toprule
\textbf{Dataset} & \textbf{Models} &  \textbf{BLEU-1} & \textbf{BLEU-2} & \textbf{BLEU-3} & \textbf{BLEU-4} & \textbf{METEOR} & \textbf{SPICE} & \textbf{ROUGE-L} & \textbf{CLIP} \\

 \midrule
\multirow{3}{*}{Artpedia} 
 & Wu2022~\cite{Wu_2022}  & 24.7 & -- & -- & 3.06 & 6.58   & -- & 22.4 & -- \\
 & KALE (w/o metadata)~\cite{ijcai2024p848}  & 29.9 & 15.0 &  7.95 & 4.77 & 8.02 & 5.49 & 22.4 & -- \\
 & KALE (w/ metadata)~\cite{ijcai2024p848}   & 32.6 & 17.7 & 10.9 & 7.48 & 9.33 & 7.68 & 23.7 & -- \\
 \rowcolor{gray!20} 
 & ArtRAG (4o-mini)     &  37.6 &  18.7 &  8.25  & 6.17 &  13.9  &  9.36 &  25.8 &  84.8\\
 \rowcolor{gray!20} 
  & ArtRAG (Qwen2VL) & \textbf{40.2}  & \textbf{19.7}  & \textbf{13.2}   &\textbf{10.86} & \textbf{14.3}  & \textbf{10.3}  & \textbf{28.3}  & \textbf{84.9}\\
\midrule
\multirow{3}{*}{SemArt} 
 & Bai~\cite{Bai2021ExplainMT}   & --  & --  & -- & \textbf{9.1} & 11.4 & -- & 23.1 & -- \\
 & KALE (w/o metadata)~\cite{ijcai2024p848}   & 25.9 & 13.8 & 8.8 & 6.7 & 7.5 & 6.00 & 19.9 & -- \\
 & KALE (w/ metadata)~\cite{ijcai2024p848}   & 27.7   & 15.7 & \textbf{10.8} & 8.6 &  9.5 & 7.3    & 21.9 & -- \\
 \rowcolor{gray!20} 
 & ArtRAG (4o-mini)      & 29.27 & 11.78  &5.39   & 3.3   &  \textbf{12.47} & \textbf{8.58} & 18.88 & 84.21 \\
 \rowcolor{gray!20} 
  & ArtRAG (Qwen2VL) & \textbf{31.78}  & \textbf{15.90}  & 8.92  & 5.2 & 10.24  & 8.03   & \textbf{24.1}  &  \textbf{84.55}\\

\bottomrule 
\end{tabular}

\end{table*}

\begin{table*}[t]
\centering
\caption{Ablation study results on Artpedia and SemArt with GPT4o-mini for its balanced ability on text and image modelling. We evaluate the impact of external knowledge: (a) no extra knowledge, (b) non-structured knowledge, (c) structured knowledge using ArtGraph, and (d) our full model with dual subgraph retrieval. Bold indicates best results per dataset.}
\small

\begin{tabular}{llccccccc}
\toprule
\textbf{Dataset} & \textbf{Model Variant} & \textbf{BLEU-1} & \textbf{BLEU-2} & \textbf{BLEU-3} & \textbf{METEOR} & \textbf{SPICE} & \textbf{ROUGE-L} & \textbf{CLIP} \\
\midrule
\multirow{4}{*}{Artpedia} 
& (a) No Extra Knowledge                & 30.6  & 13.1  & 5.5   & 9.84  & 4.92  & 18.0  & 81.71 \\
& (b) + Non-Structured Knowledge        & 34.8  & 14.1  & 6.4   & 11.1  & 5.60  & 19.1  & 79.01 \\
& (c) + Structured Knowledge (ACKG) & 36.6  & 16.7  & 7.68  & 13.7  & 8.21  & 21.8  & 83.12 \\
 \rowcolor{gray!20} 
& (d) + Multi-granularity Retriever (Full)  & \textbf{37.6}  & \textbf{18.7}  & \textbf{8.25}  & \textbf{13.9}  & \textbf{9.36}  & \textbf{25.8}  & \textbf{84.83} \\
\midrule
\multirow{4}{*}{SemArt} 
& (a) No Extra Knowledge                & 22.7  & 8.39  & 2.98  & 8.4   & 7.38  & 17.53 & 81.86 \\
& (b) + Non-Structured Knowledge        & 23.72 & 8.82  & 3.42  & 9.82  & 7.04  & 17.81 & 82.72 \\
& (c) + Structured Knowledge (ACKG) & 27.73 & 10.5  & 5.38  & 11.33 & 7.94  & 18.5  & 83.56 \\
 \rowcolor{gray!20} 
& (d) + Multi-granularity Retriever (Full)  & \textbf{29.27} & \textbf{11.78} & \textbf{5.39} & \textbf{12.47} & \textbf{8.58} & \textbf{18.88} & \textbf{84.21} \\
\bottomrule
\end{tabular}

\label{tab:ablation}
\end{table*}

\textbf{Fine-Grained Multimodal Pruning}: 
After retrieving a subgraph with the top-$k$ most relevant nodes and their edges, we apply fine-grained pruning to eliminate redundant or irrelevant content that retrieved in the first stage. This refines the subgraph into a more coherent contextual set.

\textit{Multimodal Similarity Measurement.} 
To assess the semantic alignment between each retrieved node and input painting, we introduce a relevance scoring mechanism tailored to painting-specific context. 
We then encode the painting’s metadata and visual features and compare them against the node descriptions using a MLLM. The MLLM is prompted to rank the nodes by their contextual relevance, capturing both explicit metadata links and implicit contextual associations. A multimodal similarity score $s_{\text{ms}}(v_i)$ is then derived from the ranking position to quantify each node’s $v_i$ alignment with the painting.

\textit{Graph-Based Importance Scoring:}  
To complement semantic filtering, we incorporate structural relevance using graph centrality. Specifically, we prioritize nodes with high degree centrality, which represent influential artists, movements, or techniques that act as key connectors in the ACKG. Including such nodes ensure broader contextual grounding and interpretability. We compute a centrality score $s_{\text{gc}}(v_i)$ for each node $v_i$ to quantify its structural importance.

Overall, a pruning function $\mathcal{P}$ selects the top-$m$ nodes ($m < k$) from the initially retrieved set $\mathcal{V}_k$ based on a joint scoring mechanism that combines two score of $s_{\text{gc}}(v_i)$  and $s_{\text{gc}}(v_i)$. The final node score is computed as a weighted combination with normalization:
\begin{equation}
    s(v_i) = \lambda \cdot \text{Norm} (s_{\text{ms}}(v_i)) + (1 - \lambda) \cdot \text{Norm} (s_{\text{gc}}(v_i)), \quad v_i \in \mathcal{V}_K
\end{equation}
where $\lambda \in [0, 1]$ balances semantic alignment and graph connectivity. The top-$m$ nodes ranked by $s(v_i)$ are retained, along with their inter-connected edges, forming a refined subgraph:

\begin{equation}
\mathcal{V}' = \operatorname*{arg\,top\text{-}m}_{v_i \in \mathcal{V}_k} \; s(v_i)
\end{equation}
\begin{equation}
\quad
\mathcal{E}' = \{(v_i, v_j) \in \mathcal{E}_k \mid v_i, v_j \in \mathcal{V}'\}
\end{equation}
\begin{equation}
\mathcal{G}' = (\mathcal{V}', \mathcal{E}'),
\end{equation} where $\mathcal{V}'$ and $\mathcal{E}'$ are the nodes and their connected edges after fine-grained pruning. The subgraph $\mathcal{G}'$ offers a compact, semantically relevant context to the painting. Our multi-granularity structured retriever, combining multimodal similarity and graph-based importance, ensures contextual alignment and structural coherence.

\subsection{Augmented generation with in-context learning}
The final step generates rich, multi-perspective painting descriptions using the contextual subgraph retrieved earlier. By combining visual content, metadata, and structured graph context, the model can address both content and contextual dimensions. We design in-context learning prompts that pair metadata with the subgraph, illustrating how structured inputs map to high-quality descriptions. This guides the MLLM to produce responses aligned with the user’s query and the painting’s visual and historical context. The prompt structure we used is shown in supplementary material.

\section{Experiments \& Results}
Evaluating generated text is inherently challenging, and this complexity is amplified in the case of visual painting explanations, which require a nuanced understanding of artistic, historical, and cultural contexts.
We conducted extensive experiments both objectively and subjectively
to evaluate the effectiveness of ArtRAG, focusing on the following key research questions:

\begin{enumerate}
    \item Overall Performance: How does ArtRAG perform compared to baselines in generating painting descriptions?
    
    \item Effectiveness of ACKG and Structured Retrieval: What is the impact of incorporating an ACKG and the proposed multi-stage retrieval mechanism on description quality?

    \item Multi-Perspective Contextual Understanding: How well does ArtRAG capture diverse contextual dimensions such as content, form, and cultural background?

    \item Human Evaluation: How does ArtRAG compare to baselines in terms of human-judged quality and contextual richness of the generated descriptions?
\end{enumerate}

\subsection{Experimental Setup} 

We evaluate our approach using two benchmark datasets for multi-perspective artwork description generation: SemArt and Artpedia that contain multiple painting descriptions per painting image.

The SemArt~\cite{semart_2018} dataset, originally developed for cross-modal retrieval tasks, features European fine art reproductions sourced from the Web Gallery of Art.  The dataset includes painting textual descriptions on content, form and context. The data were then refined by further annotating the explanations~\citep{Bai2021ExplainMT}.

Artpedia~\cite{2019_Artpedia} consists of paintings spanning the period from the 13th to the 21st century crawled from Wikipedia, each associated with textual descriptions and a corresponding title. This dataset provides diverse content, but requires further contextual enrichment to support multi-topic description generation.
To evaluate the performance of our approach, we compare it against three state-of-the-art baselines: Wu2022~\cite{Wu_2022}, Bai(2021)~\cite{Bai2021ExplainMT}, and KALE(2024)~\cite{ijcai2024p848}, and also use GPT4o (version 2024-08-06) for qualitative comparsion.

\begin{figure*}[h!]
    \centering
    
    \vspace{-4mm}
\includegraphics[width=\linewidth]{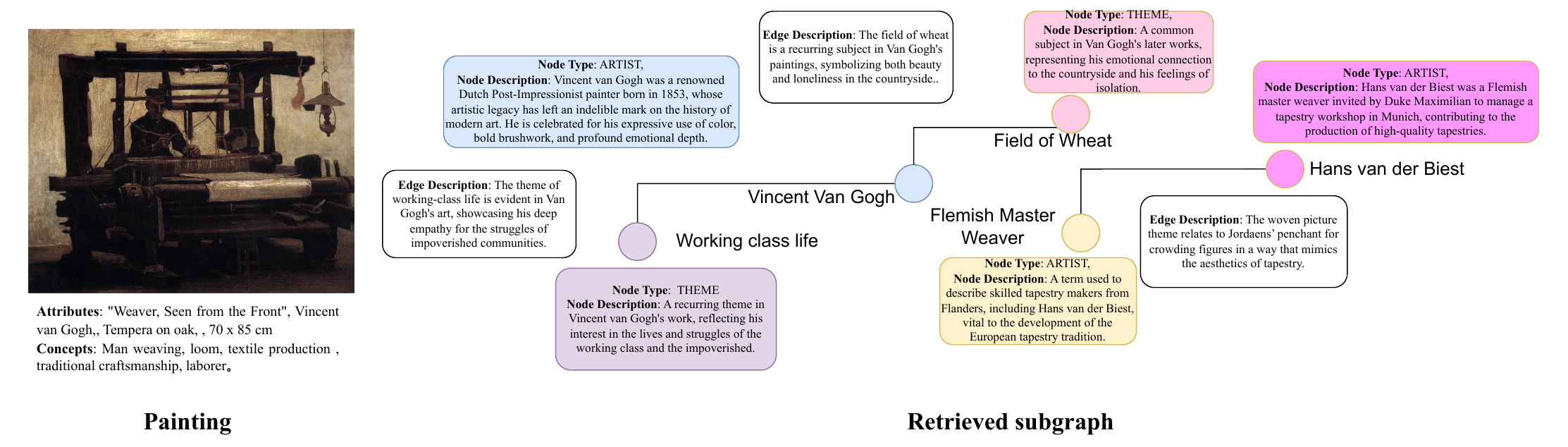}
    \vspace{-6mm}
    \caption{
    Visual example of the contextual subgraph $\mathcal{G}'$ retrieved by ArtRAG. This illustrates how relevant entities and their semantic relations are retrieved from the ACKG based on a target painting \ptitle{Weaver, Seen from the Front} by Vincent van Gogh}
    \label{fig:qual_example2}
\end{figure*}

\begin{figure*}[ht!]
    \centering
    
    \includegraphics[width=\linewidth]{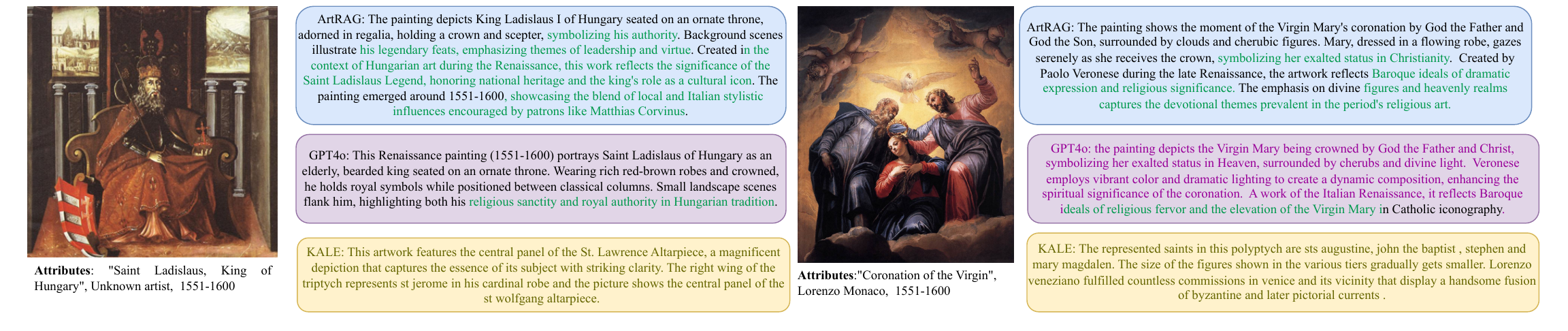}
    \vspace{-6mm}
    \caption{Qualitative comparison of painting explanations generated by ArtRAG (ours), KALE, and GPT-4o. context explanation is highlight in \textcolor{Green}{green}.}
    \vspace{-2mm}
    \label{fig:qual_example1}
\end{figure*}

 Our implementation leverages GPT-4o-mini (version: 24-07-18)~\cite{GPT-4o} for knowledge graph construction, due to its strong capabilities in both language modeling and visual understanding. For the final explanation generation, we also experimented with Qwen2-VL-7B~\cite{qwen2vl} as an alternative MLLM for its superior result in vision understanding. We detect 5 concepts for each painting. 
 In our experiments, the multi-granularity structured context retriever selects $k = 10$ nodes at the coarse level and $m = 5$ nodes at the fine-grained level. We use a softmax function for score normalization, and set the weighting score $\lambda = 0.5$ .

\subsection{Overall Performance}
 Table~\ref{tab:evaluation_metrics} presents the evaluation results of our proposed ArtRAG framework (using GPT4o-mini and Qwen2-VL~\cite{qwen2vl}) against several baseline models on the Artpedia and SemArt datasets. Baseline numbers are taken from their original publications. We report standard generation metrics, including BLEU~\cite{2002-bleu}, METEOR~\cite{2005-meteor}, SPICE~\cite{spice_2016}, ROUGE-L~\cite{2004_rouge}, and CLIPScore~\cite{2021-clipscore}, to benchmark the quality of generated descriptions. While these metrics are widely used in captioning tasks, we acknowledge that they are not ideally suited for the art domain or for evaluating context-enriched art explanations. Nonetheless, we adopt them to maintain comparability with prior work and to provide a general sense of fluency and overlap with ground-truth annotations. The highest scores for each metric are bolded for clarity.
 On the Artpedia dataset, ArtRAG achieves the best performance across most metrics, including BLEU-1 (40.2), METEOR (14.3), SPICE (10.3), and ROUGE-L (28.3), indicating improved alignment with ground-truth text. On the SemArt dataset, ArtRAG (GPT4o-mini) also leads on BLEU-1 (31.8) and METEOR (11.51), showing its ability to produce more accurate and detailed outputs even without task-specific training. We also observe ArtRAG does not achieve the highest scores on BLEU-3 and BLEU-4 on SemArt due to their strict requirement for exact 3- or 4-gram matches, which is challenging for the subjective nature of art descriptions. Our model, without task-specific fine-tuning, favors semantic richness over rigid phrasing.

\subsection{Effectiveness of ACKG and structured context Retriever}
Table 3 presents the results of our ablation study on the Artpedia and SemArt datasets, evaluating the impact of incorporating external knowledge into the caption generation process. We compare four model variants: (a) a baseline with no external knowledge (i.e., no RAG), (b) a model augmented with non-structured Wikipedia text (as in Section~4.1), (c) a model leveraging ACKG, and (d) our full model with multi-granularity retriever. Across both datasets, we observe a consistent improvement in all evaluation metrics as more structured knowledge is incorporated. 
 (a) No-RAG baseline achieves the lowest performance across all metrics, highlighting the limitations of LLMs when relying solely on internal knowledge without external contextual information. (b) Naive-RAG improves slightly over No-RAG, demonstrating the value of incorporating external knowledge. (c) shows the advantage of ACKG with naive retriever. The full model with multi-granularity structured retriever  (d) achieves the highest performance across all metrics. 

\begin{figure*}[th!]
    \centering
    
    \includegraphics[width=\linewidth]{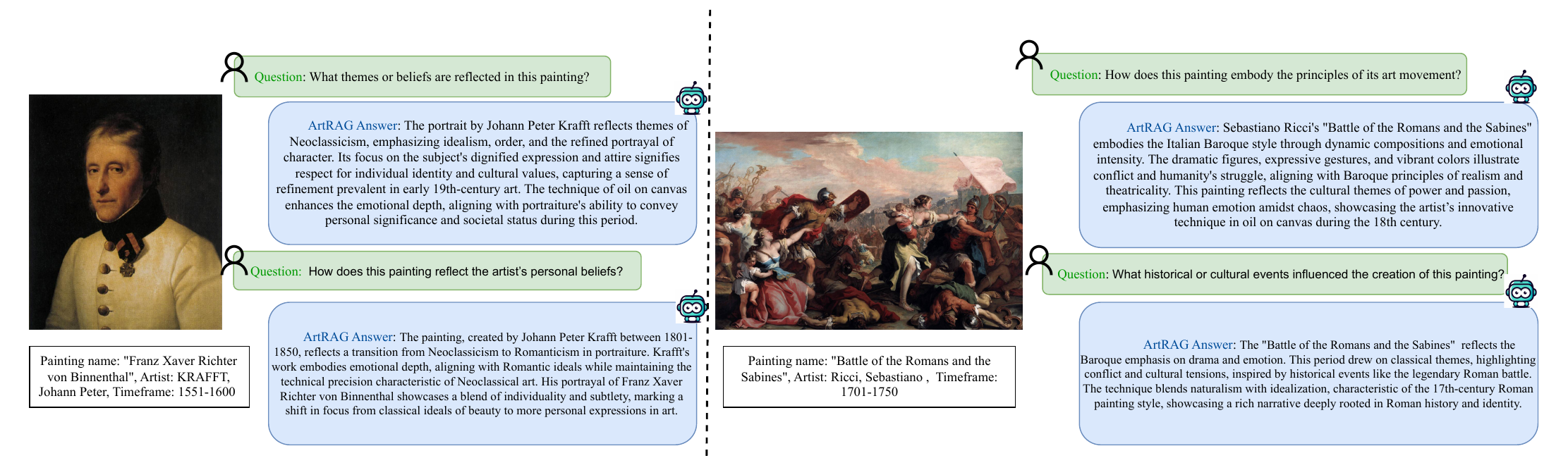}
    \vspace{-6mm}
    \caption{Qualitative examples of ArtRAG's ability to answer context-specific questions about paintings across multiple interpretive dimensions. }
    \vspace{-3mm}
    \label{fig:qual_conv_example}
\end{figure*}

\subsection{Multi-Perspective Contextual Understanding}

  We show the sub-graph retrieved by the Multi-granularity structured context Retriever. As shown in Figure~\ref{fig:qual_example2}, for painting \ptitle{Weaver, Seen from the Front} by Vincent van Gogh, the multi-granularity structured context retriever identifies and connects a set of semantically relevant nodes, including themes such as working-class life and textile production, and historical figures like Hans van der Biest and Flemish master weavers. The retrieved graph highlights both direct semantic links and indirect cultural associations, illustrating ArtRAG’s ability to assemble a rich, interpretable context.
 
 Figure~\ref{fig:qual_example1} shows example explanations generated by different models. The descriptions vary in their emphasis on historical context, artistic style, and symbolic interpretation. ArtRAG tends to integrate cultural and historical insights with stylistic analysis, while GPT-4o focuses on a more general descriptive and symbolic narrative. KALE highlights structural and compositional elements, often referencing artistic traditions. These differences illustrate the varying strengths and limitations of each approach in capturing the multifaceted nature of art interpretation.
 To further assess ArtRAG’s ability to generate meaningful descriptions, we present qualitative examples where the model answers context-specific questions about paintings in Figure~\ref{fig:qual_conv_example}. Unlike general captioning approaches, which describe visual elements in isolation, ArtRAG generates detailed, multi-perspective explanations that integrate knowledge of art history, artistic movements, and cultural context. In the first example, ArtRAG identifies Neoclassical and Romantic influences in Franz Xaver Richter von Brünnigk’s portrait and contextualizes how the artist’s stylistic transition reflects evolving artistic values of the time. Similarly, in Battle of the Romans and the Sabines, ArtRAG accurately connects the composition’s dramatic intensity and structured dynamism to Baroque painting techniques, highlighting its ties to historical traditions and artistic innovation. These results suggest that ArtRAG not only captures visual details but also understands the broader interpretive context of artworks, an essential capability for AI-assisted art interpretation. The model generalizes well across different question types, reinforcing its potential for interactive and explainable AI applications in digital humanities and cultural studies.

\begin{table}[h!]
    \centering
    \caption{Human evaluation ranking results for each model. We report the percentage of times each model was ranked 1st, 2nd, or 3rd by annotators, along with the average rank. }
    \begin{tabular}{lcccc}
        \toprule
         \textbf{Model}  & \textbf{Rank 1st} & \textbf{Rank 2nd} & \textbf{Rank 3 rd} &\textbf{Avg Rank}\\
        \midrule
        KALE &  19.8\%     & 22.9\%    & 57.4\% &2.38 \\
        GPT4o\footnote{GPT4o-2024-0806}    & 28.3\%      & 39.0\%     & 32.6\% & 2.04\\
        ArtRAG   & 51.9\%     & 38.1\%     &  10.0\%    & \textbf{1.58}\\
        \bottomrule
    \end{tabular}

    \label{tab:Human_eval}
\end{table}

\subsection{Human evaluation}
We conduct a human evaluation to assess the quality of descriptions generated by different models. We randomly select 60 paintings from the test dataset and present their metadata (title, artist, and creation year), along with three descriptions generated by ArtRAG, KALE, and GPT-4o, to a group of 10 evaluators. Each painting is evaluated by at least three different evaluators. Given that art evaluation is inherently subjective and requires domain expertise, we recruit human experts with backgrounds in the cultural industry, multimedia and innovation research. Evaluators are asked to rank the three descriptions for each painting in order of overall quality (1st, 2nd, 3rd). The evaluation criteria focus on the following aspects:
Relevance: How well does the description correspond to contextual elements of the painting? Context Presence: Does the explanation reflect historical, cultural, or thematic background?
Comprehensiveness: Does the description go beyond surface-level analysis to capture multi-dimensional interpretation?

\begin{figure}[h!]
    \centering
    
    \includegraphics[width=\linewidth]{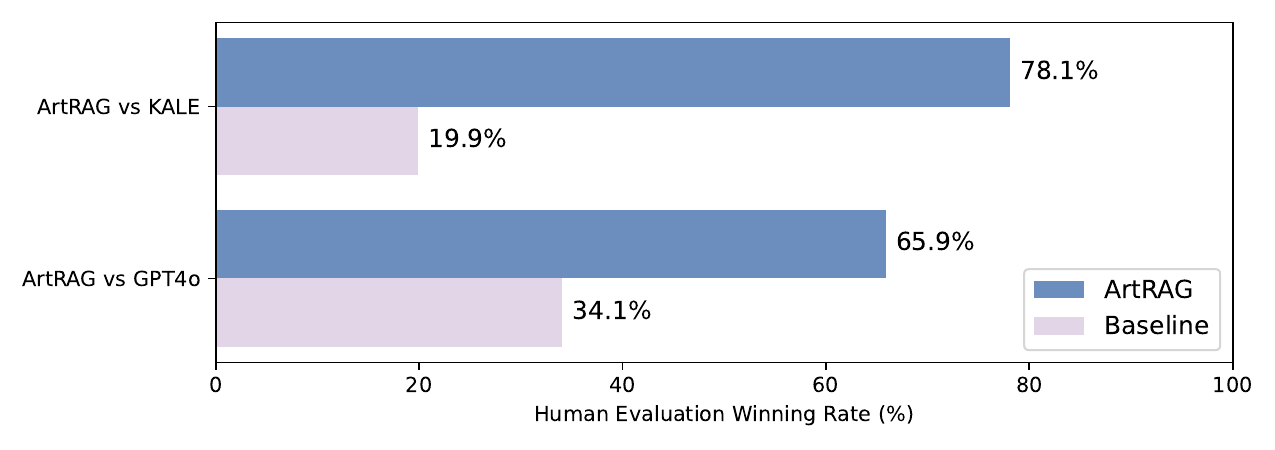}
    \vspace{-5mm}
    \caption{ Pairwise winning on human evaluation of ArtRAG over GPT4o and KALE model}
    \vspace{-3mm}
    \label{fig:winning_rate}
\end{figure}

The results are summarized in Table~\ref{tab:Human_eval}. ArtRAG significantly outperforms both KALE and GPT-4o, with 51.2\% of its outputs ranked 1st and an average rank of 1.58, compared to 2.04 for GPT-4o and 2.38 for KALE. In addition to the ranking-based comparison, Figure~\ref{fig:winning_rate}  presents pairwise win rates, showing that ArtRAG outperforms GPT-4o in 65.9\% and KALE in 78.1\% of cases. These results highlight ArtRAG's strength in generating more coherent, relevant, and contextually enriched painting descriptions.

\section{Conclusion and future work}
We present ArtRAG, a novel, training-free framework for visual art understanding. ArtRAG uses a structured, domain-specific Art Context Graph to integrate multi-perspective knowledge—from visual content to historical context—into a retrieval-augmented generation process. This allows it to create informative and interpretable art explanations without task-specific fine-tuning. Our evaluations show that ArtRAG goes beyond simple captioning, providing insights that align with art context reasoning. We also acknowledge the inherent challenge of evaluating subjective artistic explanations and suggest incorporating expert perspectives for more consistent evaluation criteria.

\section{Acknowledge}
We extend our sincere gratitude to the University of Amsterdam Data Science Center for their  support of this research. We also deeply appreciate our colleagues from the University of Amsterdam for their insightful discussions and contributions to the human evaluation.
\bibliographystyle{ACM-Reference-Format}
\bibliography{references}

\appendix

\section{Supplementary Material}

Due to space constraints in the main paper, we omit certain implementation details—such as full prompting strategies used for ACKG building—and limit some sections to a single illustrative example. In this supplementary material, we first provide the complete prompting details, followed by additional qualitative examples to offer a more comprehensive and convincing analysis of our method.

\begin{figure*}[h]
    \centering\includegraphics[width=0.9\linewidth]{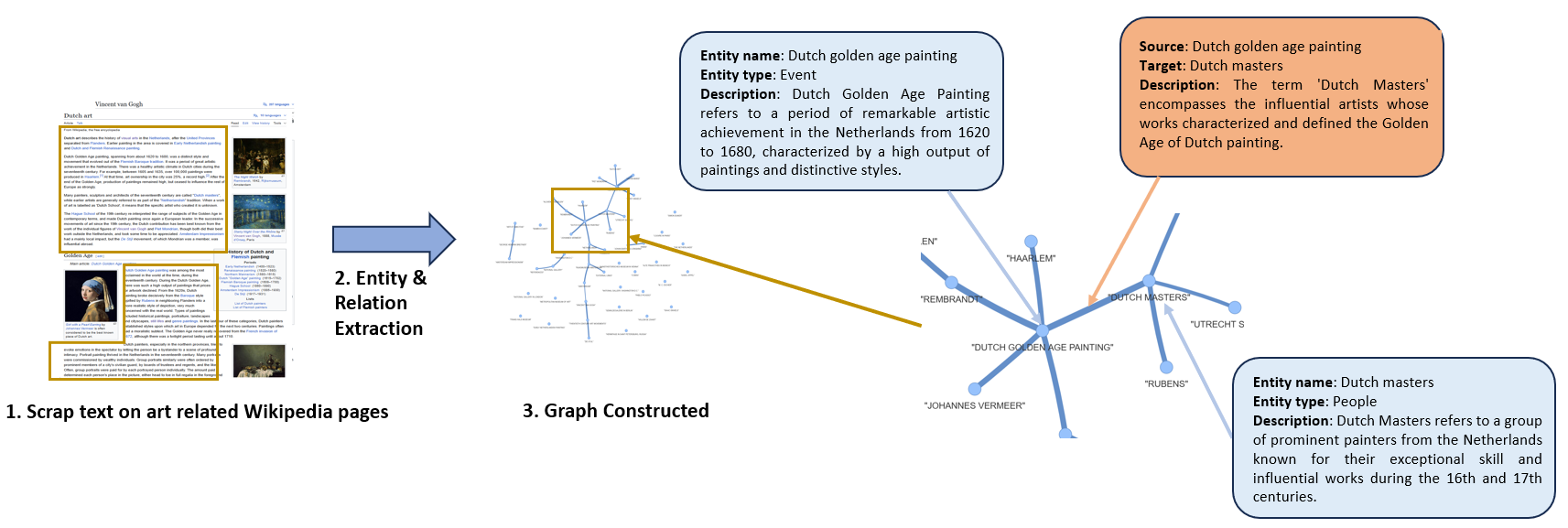}
    \vspace{-3mm}
    \caption{Example of automatic entity and relation extraction from Wikipedia text for Art Context Knowledge Graph (ACKG) construction. Starting from a scraped Wikipedia page (e.g., “Dutch art”), our framework extracts key entities such as Dutch golden age painting (Culture \& History), Dutch masters (Artist), and associated descriptions. It also identifies semantic relationships, such as “Dutch masters” being influential figures in the “Dutch golden age painting” era. The resulting nodes and edges are used to construct a structured graph capturing the interconnected context of art history.}
    \label{fig:graph_example}
\end{figure*}

\subsection{ACKG construction details}

We scraped over 1,000 Wikipedia pages shown in Tab.\ref{tab:wiki_stats} related to art, guided by category tags and entity types from SemArt, ArtPedia, and ExpArt. Each page was segmented into overlapping text chunks using a sliding window (1000 tokens with 100-token stride), enabling finer-grained concept and relation extraction. in Fig.\ref{fig:graph_example}, we show a example of A subgraph snapshot built from Wikipedia page "Dutch art".

Fuzzy Matching for Node Deduplication:
To reduce redundancy in the constructed graph, we perform post-processing to merge nodes with similar surface forms or semantically equivalent meanings. We compute the pairwise string similarity between node names using normalized Levenshtein distance, implemented via the RapidFuzz library. Two nodes are merged if their similarity exceeds a threshold of 0.95 and their types are compatible (e.g., both are "Art Movement" nodes).

In below, we show the prompt we used to extract nodes and edges from wikipedia documents:

\begin{table}[h]
    \centering
    \caption{Statistics of collected Wikipedia pages and word counts for different art-related categories.}
    \begin{tabular}{lccc}
        \toprule
         \textbf{Document Type} & \textbf{ Pages} & \textbf{Word number} & \textbf{Token number}\\
        \midrule
        Artists         & 952     & 1750K   & 2067K\\
        Art schools      & 20      & 69K     & 80K \\
        Art Types        & 9       & 37K     &  43K \\
        Cultural events & 85      & 781K    &  895K\\
        Art movements   & 25      & 128K    &  159K  \\
        \midrule
        Total           & 1091    & 2765K  & 3244K\\
        \bottomrule
    \end{tabular}

    \label{tab:wiki_stats}
\end{table}

\begin{tcolorbox}[breakable, fontupper=\customfont, title=Prompt for Node and edge extraction]
{\small
\textbf{Input Data}: Art related Document chunk\\
\textbf{Instruction}: 
Given a visual art related text document that is potentially relevant to this activity and a list of entity types, identify all entities of those types from the text and all relationships among the identified entities.

\textbf{Steps}:
1. Identify all entities. For each identified entity, extract the following information: \\
- entity\_name: Name of the entity, capitalized \\
- entity\_type: One of the following types: [{entity\_types}] \\
- entity\_description: Comprehensive description of the entity's attributes and activities

2. From the entities identified in step 1, identify all pairs of (source\_entity, target\_entity) that are *clearly related* to each other. \\
For each pair of related entities, extract the following information: \\ 
- source\_entity: name of the source entity, as identified in step 1 \\ 
- target\_entity: name of the target entity, as identified in step 1 \\
- relationship\_description:

------ Example -----

\textit{Contextual Graph Snapshot:}
\begin{itemize}
  \item \textbf{Nodes:}....
  \item \textbf{Edges:}....
\end{itemize}

}
\end{tcolorbox}

\begin{table*}[h]
\centering 
\caption{Illustrative examples of node types and representative entities in the Art Context Knowledge Graph.}
\label{tab:node_examples}
\begin{tabular}{ll}
\toprule
\textbf{Node Type}         & \textbf{Example Entities} \\
\midrule
\textbf{Artist}                    & Paul Cézanne, Vincent van Gogh, Claude Monet, Jacques-Louis David \\
\textbf{Theme}                     & Working-class life, Spiritual devotion, Balance and Harmony \\
\textbf{Culture \& History}        & Flemish Renaissance, Dutch Golden Age, Christian faith  \\
\textbf{Art Style \& Technique}    & Oil on canvas, Fresco, Chiaroscuro, Linear perspective \\
\textbf{Art Movement \& School}    & Impressionism, Surrealism, Bauhaus, Florentine School \\
\bottomrule
\end{tabular}
\end{table*}

\begin{figure*}[h]
    \centering\includegraphics[width=0.9\linewidth]{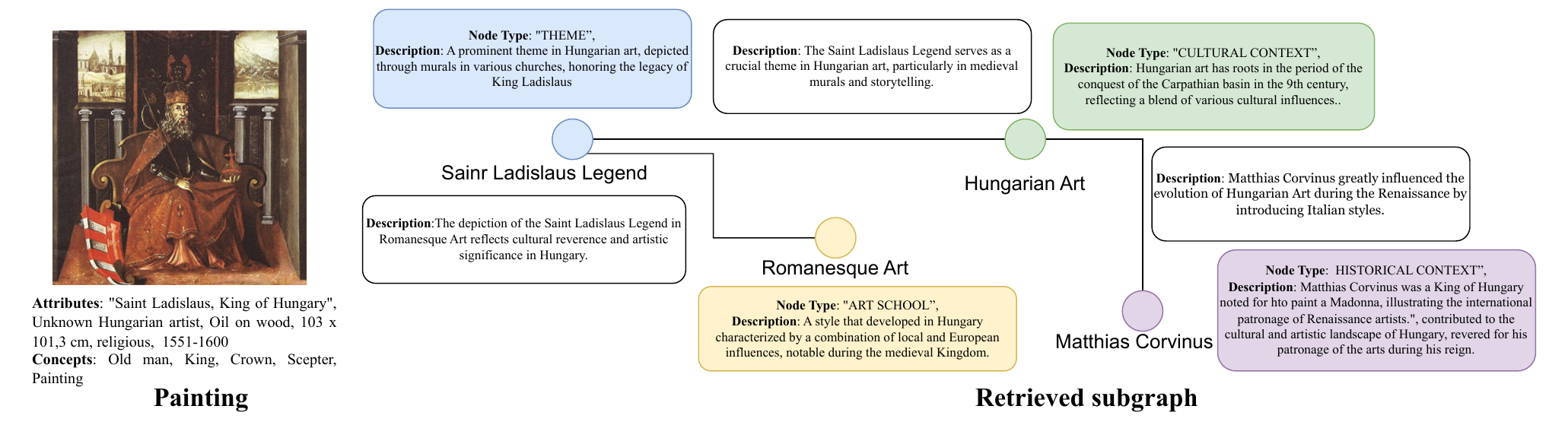}
    \vspace{-3mm}
    \caption{Visual example of the contextual subgraph $\mathcal{G}'$ retrieved by ArtRAG for painting \ptitle{Saint Ladislaus, King of Hungary.}}
    \label{fig:graph_example-2}
\end{figure*}

\subsection{Concept detection}
To support downstream context retrieval and graph grounding, we extract high-level visual and thematic concepts from each painting. We design a dedicated prompt that instructs a multimodal language model to analyze both the visual features of the painting and its associated metadata (e.g., title, artist, creation period) to generate a short list of relevant concepts, as shown bwlow:
\begin{tcolorbox}[breakable, fontupper=\customfont, title=Prompt for concept detection]
{\small
\textbf{Input Data}:  Painting attributes, and visual painting \\
\textbf{Instruction}: 
You are an expert in art interpretation. Given an image of a painting along with its title, artist, and creation period, identify a list of high-level concepts that capture the key concptes of the painting.

\textbf{\color{black}{Formatting and Constraints}}: 
Output should be formatted in Markdown. If information is missing, do not hallucinate or fabricate details.
}
\end{tcolorbox}

\begin{figure*}
\centering\includegraphics[width=0.9\linewidth]{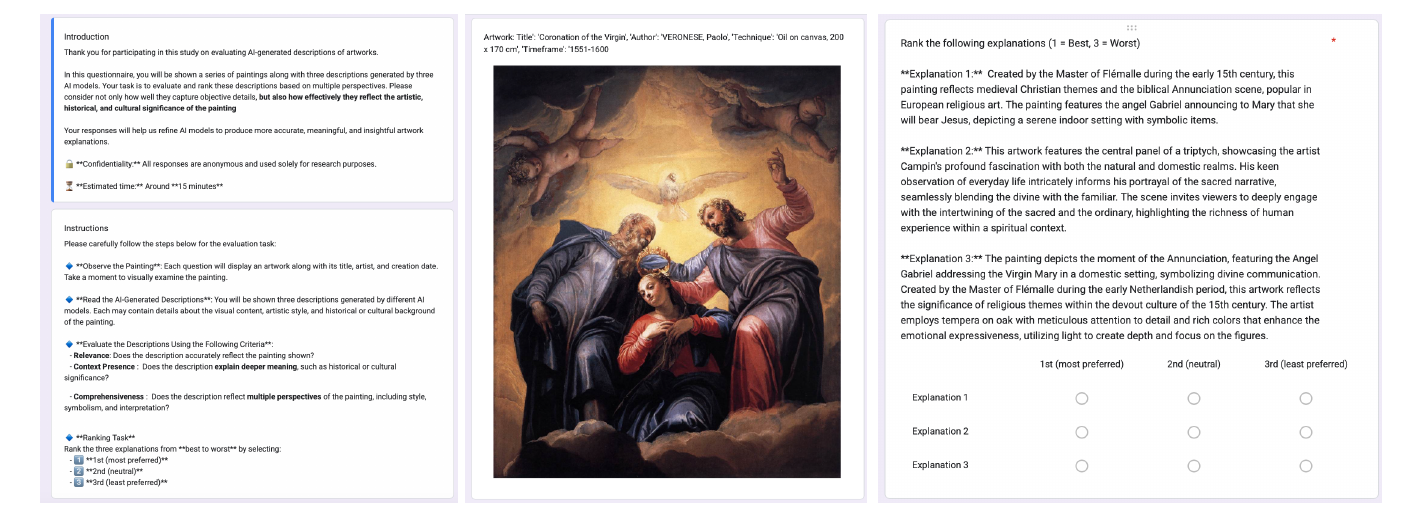}
    \vspace{-3mm}
    \caption{ Human evaluation form design, containing the description of the task, the painting information, and three random shuffled result.}
    \label{fig:human_eval_example}
\end{figure*}

\subsection{Human evaluation details}
Figure \ref{fig:human_eval_example} shows the human evaluation interface used in our study. Each task instance presents annotators with a painting image, associated metadata (title, artist, time period), and three model generated descriptions, shown in randomized order. Participants are asked to assess and rank the explanations from best to worst based on criteria such as clarity, relevance, and contextual depth. The interface also includes detailed instructions to guide the annotators and ensure consistent evaluations across participants.

\begin{figure*}[ht]
    \centering\includegraphics[width=0.8\linewidth]{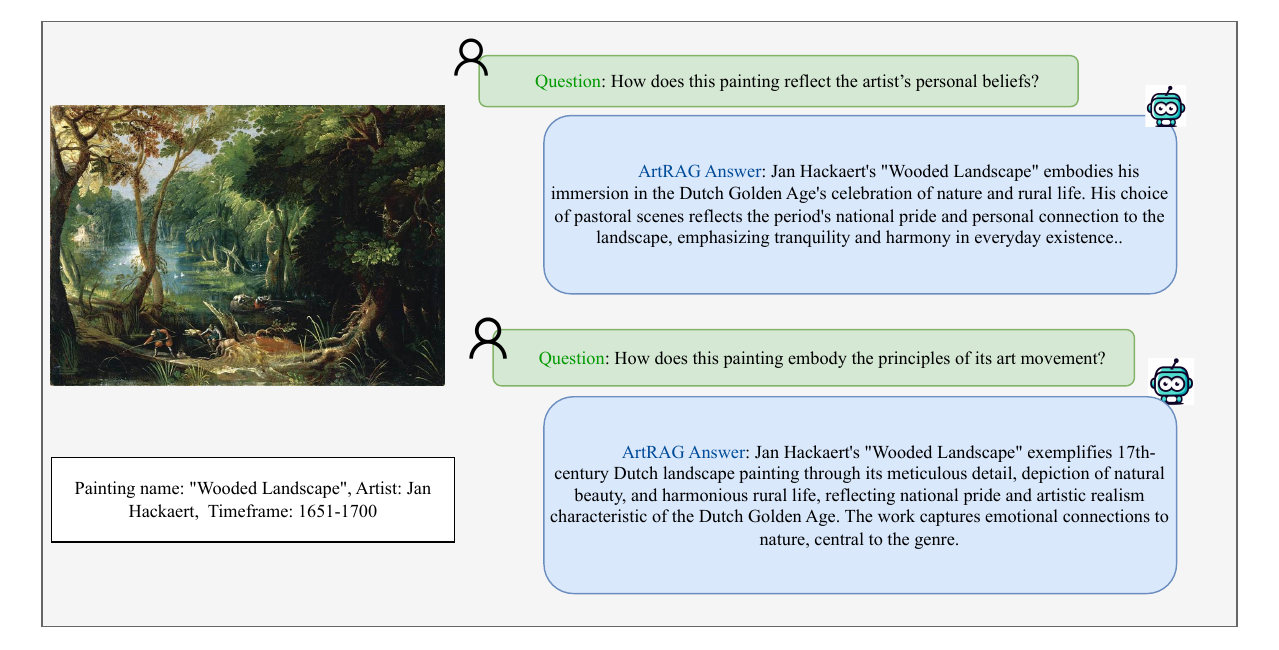}
    \caption{ Qualitative examples of ArtRAG answering multi-perspective questions about paintings. The painting “Wooded Landscape” by Jan Hackaert is interpreted from two contextual angles: personal beliefs and artistic style. ArtRAG integrates cultural and stylistic signals—such as Dutch Golden Age pastoral themes and landscape harmony}
\end{figure*}

\subsection{More qualitative examples}
We provide additional qualitative examples that showcase the effectiveness of ArtRAG in generating rich and context-aware explanations of visual artworks. 
In Each example includes the painting image, its associated metadata, and descriptions generated by three different models: ArtRAG, GPT-4o, and KALE. These examples illustrate how ArtRAG integrates visual and contextual knowledge to produce more informative, interpretable, and historically grounded outputs. 
In particular, we highlight passages that reflect nuanced understanding of artistic style, cultural significance, or historical references, demonstrating the benefits of structured context retrieval.

\begin{figure*}[ht]
    \centering\includegraphics[width=\linewidth]{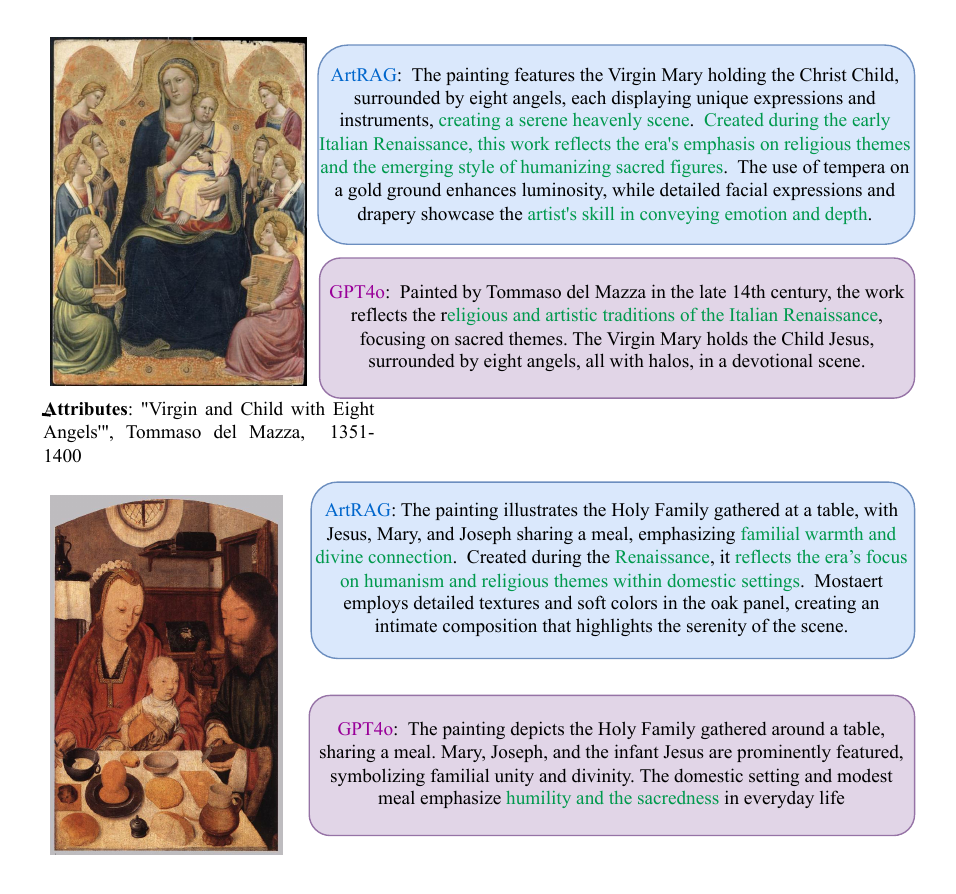}
    \caption{ Qualitative comparison of painting explanations by ArtRAG and GPT-4o on religious-themed artworks. ArtRAG captures the symbolic and stylistic nuances of the Renaissance period, interpreting how religious scenes reflect broader humanistic and theological trends. For instance, in “Virgin and Child with Eight Angels”, ArtRAG identifies the emphasis on emotional expression, sacred iconography, and evolving artistic techniques in early Italian Renaissance art. Similarly, in “Holy Family at the Table”, ArtRAG contextualizes the scene within domestic religious themes and Renaissance values. Compared to GPT-4o, which provides factual summaries, ArtRAG demonstrates deeper interpretive reasoning and awareness of historical context.}
\end{figure*}

\begin{figure*}[ht]
    \centering\includegraphics[width=\linewidth]{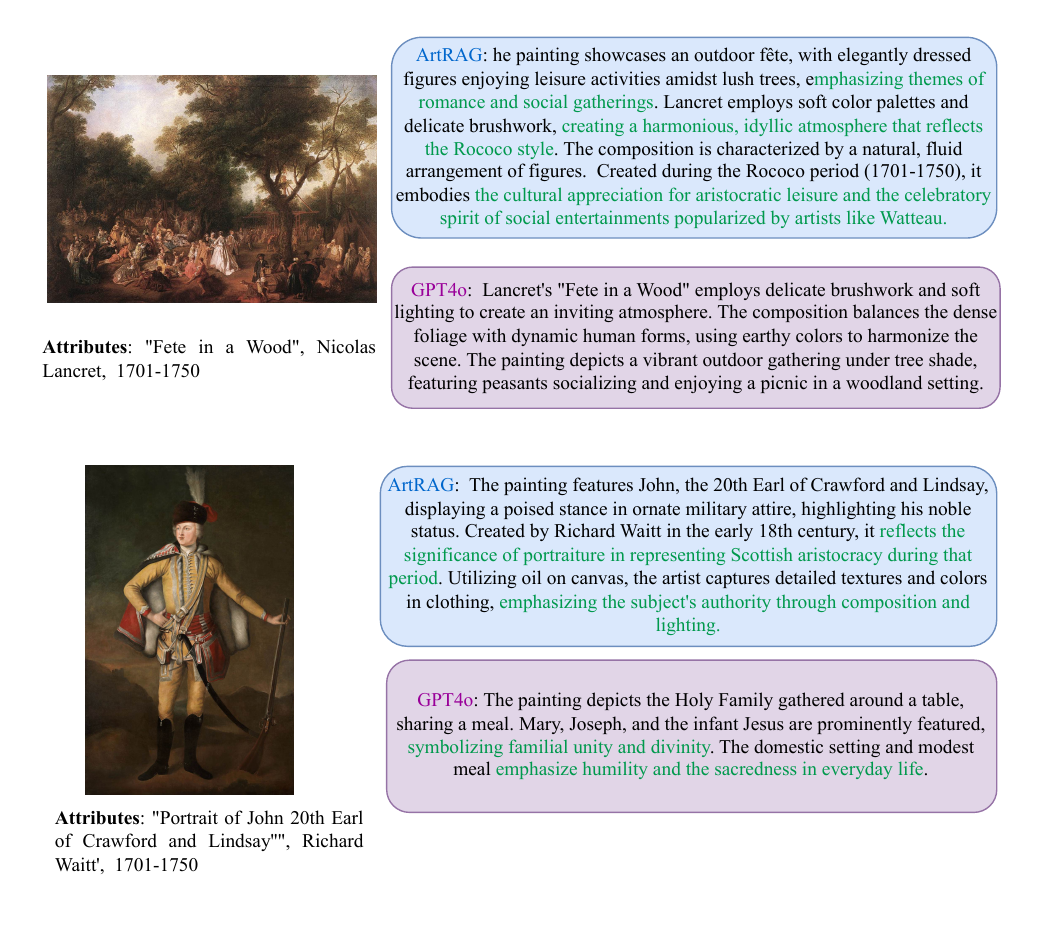}
    \caption{ Qualitative comparison of explanation generation between ArtRAG and GPT4o.For “Fête in a Wood” by Nicolas Lancret, ArtRAG highlights Rococo-era social customs, aesthetic values, and cultural influences, offering contextual richness beyond surface-level visual elements. In contrast, GPT4o focuses on scene composition without deeper historical framing. In “Portrait of John 20th Earl of Crawford and Lindsay”, ArtRAG identifies the significance of aristocratic representation in early 18th-century Scottish portraiture, connecting visual elements to socio-political context. GPT4o, however, produces a hallucinated interpretation unrelated to the painting, underscoring the importance of context-aware generation. }
\end{figure*}


\begin{figure*}[ht]
    \centering\includegraphics[width=\linewidth]{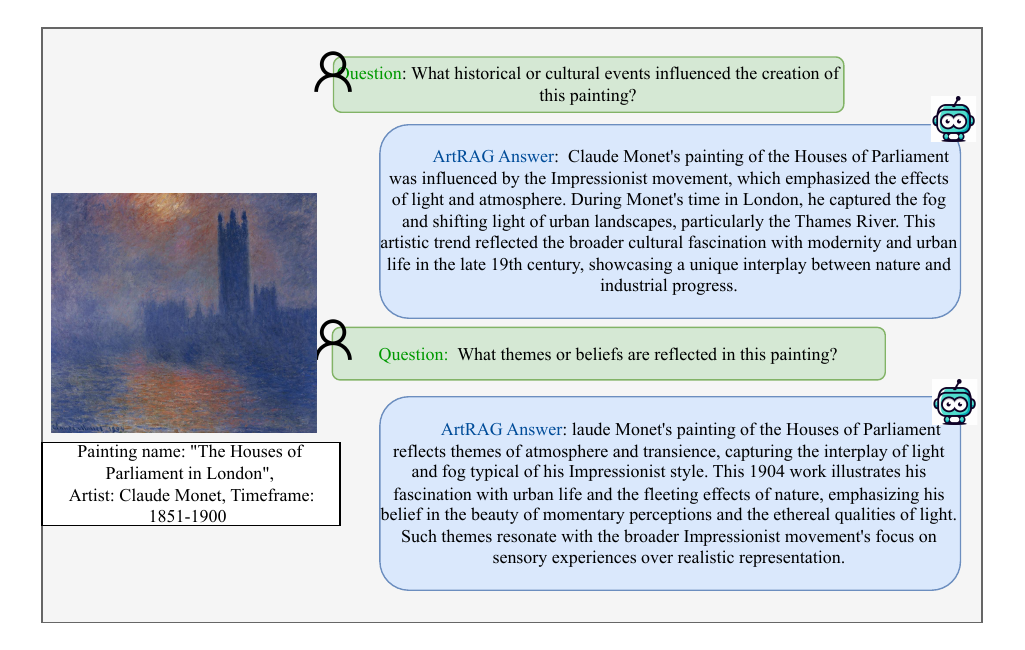}
    \caption{Human evaluation example illustrating ArtRAG’s ability to interpret paintings through historical and thematic lenses. For “The Houses of Parliament in London” by Claude Monet, ArtRAG captures how the painting reflects Impressionist ideals and cultural context—highlighting the foggy urban landscapes of 19th-century London and the artist's engagement with light and atmosphere. It also conveys the philosophical themes of transience and fleeting perception, linking visual elements to Monet’s broader artistic intent and the Impressionist movement’s ethos.}
\end{figure*}

\begin{figure*}[h]
    \centering\includegraphics[width=\linewidth]{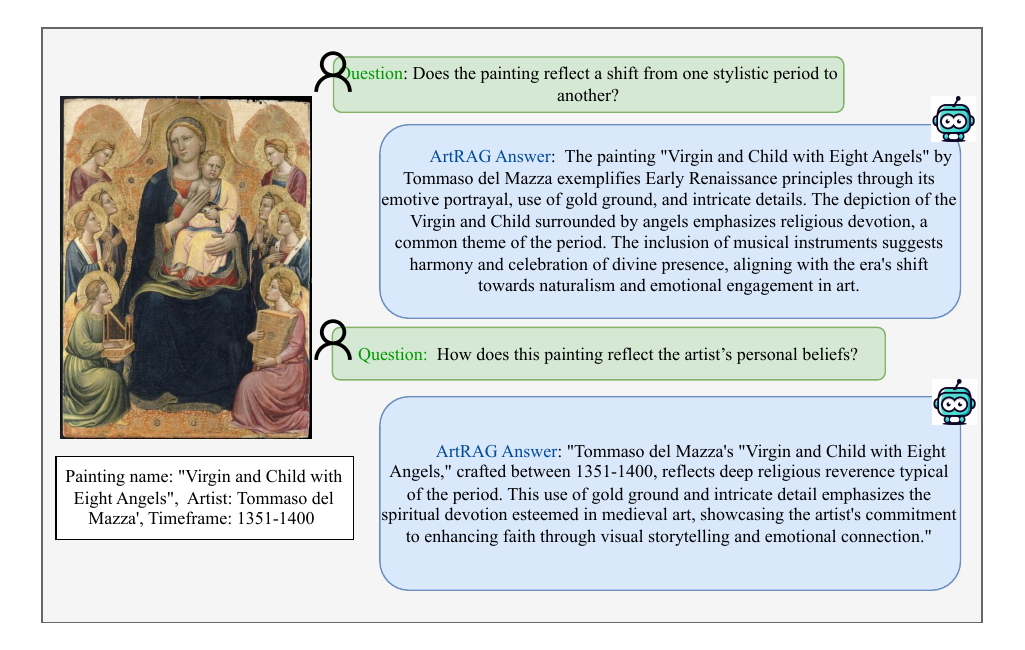}
    \caption{ Qualitative example highlighting ArtRAG’s ability to explain stylistic and thematic transitions. For “Virgin and Child with Eight Angels” by Tommaso del Mazza, ArtRAG connects religious symbolism, stylistic markers like gold background, and emotional tone to contextual shifts from medieval to early Renaissance art. These answers reflect a deep understanding of period-specific artistic values and personal devotional expression.}
\end{figure*}

\appendix

\end{document}